\documentclass[11pt,a4paper]{article}
\usepackage[hyperref]{acl2021}
\usepackage{times}
\usepackage{booktabs}
\usepackage{todonotes}
\usepackage{latexsym}
\usepackage{amsmath}

\usepackage{diagbox}
\usepackage{graphicx}
\usepackage{amssymb}

\usepackage{microtype}

\aclfinalcopy

\title{Fixing the Perspective: A Critical Examination of Zero-1-to-3}

\author{
  Jack Yu* \hspace{2em} Xueying Jia* \hspace{2em} Charlie Sun* \hspace{2em} Prince Wang \\
  \texttt{\{haoxian2, xjia2, cssun, zizhuanw\}@andrew.cmu.edu} \\
  {\footnotesize *Equal contribution}
}

\date{}

\begin{document}
\maketitle
\begin{abstract}
Novel view synthesis is a fundamental challenge in image-to-3D generation, requiring the generation of target view images from a set of conditioning images and their relative poses. While recent approaches like Zero-1-to-3 have demonstrated promising results using conditional latent diffusion models, they face significant challenges in generating consistent and accurate novel views, particularly when handling multiple conditioning images. In this work, we conduct a thorough investigation of Zero-1-to-3's cross-attention mechanism within the Spatial Transformer of the diffusion 2D-conditional UNet. Our analysis reveals a critical discrepancy between Zero-1-to-3's theoretical framework and its implementation, specifically in the processing of image-conditional context. We propose two significant improvements: (1) a corrected implementation that enables effective utilization of the cross-attention mechanism, and (2) an enhanced architecture that can leverage multiple conditional views simultaneously. Our theoretical analysis and preliminary results suggest potential improvements in novel view synthesis consistency and accuracy.
\end{abstract}

\section{Introduction and Problem Definition}
Generative models have revolutionized the creation of realistic data samples across multiple domains. Recent advances in deep learning-based approaches, particularly Generative Adversarial Networks (GANs)~\cite{goodfellow2014generative}, Variational Autoencoders (VAEs)~\cite{kingma2013auto}, and diffusion models~\cite{ho2020denoising}, have achieved remarkable success in generating high-fidelity content. These models effectively learn to capture complex data distributions, enabling the generation of novel samples that maintain perceptual and semantic consistency with training data.

The emergence of Latent Diffusion Models (LDMs)~\cite{rombach2022high} and Guided Diffusion~\cite{ho2022classifier} has particularly advanced the state of image generation. By performing the diffusion process in a learned latent space, these models achieve both computational efficiency and high-quality output. This architectural innovation has enabled significant improvements in both generation quality and control.

\paragraph{Text-to-Image Generation}
Recent breakthroughs in text-to-image generation have demonstrated unprecedented capabilities in translating natural language descriptions into visual content. Models such as DALL-E~\cite{ramesh2022hierarchical} and Stable Diffusion~\cite{rombach2022high} combine large-scale language models with diffusion-based generators to achieve remarkable photorealism and semantic alignment between textual prompts and generated images.

\begin{figure}
    \centering
    \includegraphics[width=1\linewidth]{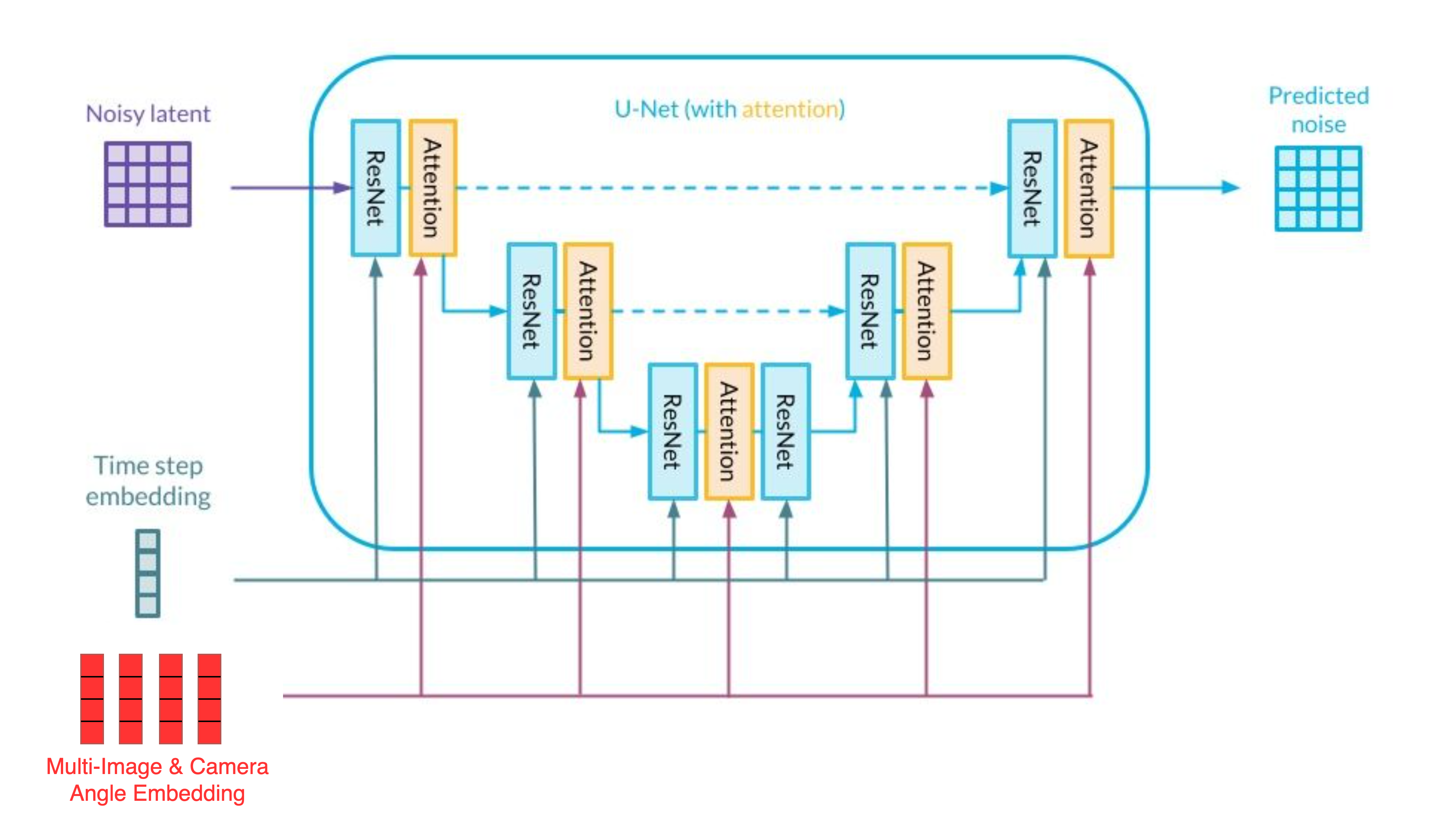}
    \caption{Architectural overview of multi-view conditional generation. The model processes multiple input views and their corresponding camera poses to generate novel viewpoints.}
    \label{fig:multi-view-condition}
\end{figure}

\paragraph{Extension to 3D Objects}
The success of text-to-image systems has catalyzed research into 3D content generation, with potential applications spanning computer-aided design, virtual reality, and robotics. However, 3D generation presents unique challenges in both representation and synthesis. Traditional 3D representations include point clouds~\cite{guo2020deep}, meshes~\cite{wang2022survey}, and voxels~\cite{blinn2005pixel}, each offering distinct trade-offs between fidelity, efficiency, and ease of manipulation. Point clouds provide geometric information but lack surface details, meshes offer complete surface descriptions but can be complex to manipulate, and voxels provide regular structure but face resolution and memory constraints.

\paragraph{Recent Advancement in 3D Generation}
Neural implicit representations, particularly Neural Radiance Fields (NeRF)~\cite{gao2022nerf}, have emerged as a promising approach for representing and rendering 3D scenes. NeRF represents scenes as continuous functions mapping 3D coordinates and view directions to color and density values. This differentiable representation enables high-quality novel view synthesis and integration with optimization-based generation approaches.

Building on these advances, several frameworks have tackled text-to-3D generation. Early approaches like Text2Shape~\cite{chen2019text2shape} established joint language-shape embeddings. More recent work, including DreamFields~\cite{jain2022zero} and DreamFusion~\cite{poole2022dreamfusion}, combines NeRF with pre-trained vision-language models like CLIP~\cite{radford2021learning} and diffusion generators to enable direct 3D synthesis from text descriptions.

\paragraph{Problem Definition}
A core challenge in 3D generation is Novel View Synthesis: generating target view images from a set of conditioning images and their relative camera poses. This capability is essential for applications ranging from 3D reconstruction to virtual reality. While Zero-1-to-3~\cite{liu2023zero1to3} has demonstrated promising results using conditional latent diffusion, it faces significant challenges in generating consistent and accurate novel views, particularly with multiple conditioning images.

\paragraph{Contributions} Our work addresses these limitations through the following contributions:

\begin{itemize}
    \item We present a detailed analysis of Zero-1-to-3's cross-attention mechanism within its diffusion UNet, revealing a critical discrepancy between the theoretical framework and implementation that impacts the processing of image-conditional context.
    \item We propose an improved architecture that corrects the identified cross-attention limitations and enables effective utilization of multiple conditional views within the UNet.
    \item Our implementation study provides insights into the practical challenges of conditional diffusion models for 3D synthesis, supported by experiments on established baselines.
\end{itemize}

\section{Related Work and Background}

\subsection{Core Technologies}
The emergence of text-to-3D generation has been enabled by several key technological advances. Neural Radiance Fields (NeRF)~\cite{gao2022nerf,mildenhall2021nerf} introduced a paradigm shift in 3D scene representation by encoding scenes as continuous volumetric radiance fields. This representation enables high-fidelity view synthesis from arbitrary camera poses while eliminating the need for extensive paired 3D datasets.

Diffusion models~\cite{ho2020denoising, rombach2022high} and score-based generative models~\cite{song2020score} have established new standards in high-quality image generation. These approaches learn to reverse a gradual noising process, demonstrating remarkable capability in generating coherent and detailed images. Their success has been particularly notable in conditional generation tasks.

The CLIP~\cite{radford2021learning} framework introduced robust joint embedding spaces for text and images through contrastive learning on large-scale web data. Combined with classifier-free guidance~\cite{ho2022classifier} and extensive image-text paired datasets, CLIP has enabled sophisticated text-controlled image generation systems~\cite{rombach2022high, ramesh2022hierarchical, nichol2021glide}.

\subsection{Text-to-3D Generation}
Recent approaches have synthesized these technologies for text-guided 3D content creation. DreamFields~\cite{jain2022zero} pioneered the optimization of NeRF representations using CLIP-based losses to align generated 3D content with text descriptions. CLIP-Forge~\cite{sanghi2022clip} and CLIP-NeRF~\cite{wang2022clip} further developed methods for leveraging CLIP embeddings in 3D shape generation and manipulation.

DreamFusion~\cite{poole2022dreamfusion} marked a significant advance by replacing CLIP-based optimization with losses derived from pre-trained text-to-image diffusion models, achieving notable improvements in text-to-3D synthesis. Magic3D~\cite{lin2023magic3d} and CLIP-Mesh~\cite{mohammad2022clip} extended these principles to optimize different 3D representations while maintaining high-quality output.

\subsection{Recent Advances in 3D Generation}
Several frameworks have recently enhanced the efficiency and quality of text-to-3D generation. DITTO-NeRF~\cite{seo2023ditto} introduced progressive 3D reconstruction, propagating high-quality information across scales for faster convergence. 3DFuse~\cite{seo2023let} integrated explicit 3D awareness into pre-trained diffusion models to improve cross-view consistency.

Zero-1-to-3~\cite{liu2023zero1to3}, which serves as our primary baseline, introduced camera perspective control in diffusion models for zero-shot 3D reconstruction from single images. Zero-1-to-3++~\cite{shi2023zero123++} enhanced this approach through:
\begin{itemize}
    \item Curated training data from Objaverse
    \item Integrated elevation conditioning
    \item Optimized training pipeline achieving 40x speedup
\end{itemize}

Magic123~\cite{qian2023magic123} represents the current state-of-the-art in single-image-to-3D generation, combining:
\begin{itemize}
    \item Dual 2D and 3D diffusion priors
    \item Progressive refinement from coarse NeRF to detailed mesh
    \item Enhanced cross-view consistency mechanisms
\end{itemize}

\subsection{Relevant Datasets}
\begin{enumerate}
    \item \textbf{Objaverse}: A comprehensive 3D dataset containing over 10 million objects with annotations, serving as the primary dataset for modern 3D generation systems.
    
    \item \textbf{NeRF4}: A balanced dataset comprising:
    \begin{itemize}
        \item 1.5GB real rendered images
        \item 1.5GB synthetic rendered images
        \item 8 scenes with 100 training, 100 validation, and 200 test images (800x800 pixels)
    \end{itemize}
    
    \item \textbf{RealFusion15}: A focused dataset containing 15 distinct objects, valuable for targeted evaluation despite its limited scale.
\end{enumerate}

\section{Baselines}

\subsection{Unimodal Baselines}
Stable Diffusion~\cite{rombach2022high}, released by StabilityAI in 2022, serves as our primary 2D generation baseline. Trained on $512 \times 512$ images from a subset of LAION-5B Dataset, its pretrained unconditional model provides a reference point for image generation quality.

DiT-3D~\cite{mo2023dit3d} explores the adaptation of diffusion transformers to 3D shape generation. The model processes point cloud data through a point-voxel transformation, where N noisy points $P_i \in \mathbb{R}^{N \times 3}$ are mapped to a point-voxel space $\mathbb{R}^{V \times V \times V \times 3}$. This representation is then split into $L = (V/p)^3$ patch tokens $t \in \mathbb{R}^{L \times 3}$ before transformer processing, enabling efficient handling of the expanded dimensionality of 3D space.

\subsection{Multimodal Baselines}
Zero-1-to-3~\cite{liu2023zero1to3} introduces camera perspective control in large-scale diffusion models, enabling zero-shot novel view synthesis and 3D reconstruction from single images. This approach forms our primary baseline for comparison.

Stable Zero-1-to-3 advances the original framework through several key improvements: enhanced training data curation from Objaverse, integrated elevation conditioning during training and inference, and an optimized training pipeline. These modifications yield a 40x speedup compared to Zero-1-to-3-XL.

Magic123~\cite{qian2023magic123} represents the current state-of-the-art in single-image-to-3D generation. The framework combines 2D and 3D diffusion priors, progressing from initial neural radiance fields for coarse geometry to memory-efficient meshes for detailed textures. Its balanced approach to exploration and precision, coupled with enhanced cross-view consistency mechanisms, demonstrates superior performance across various evaluation metrics.

RealFusion~\cite{melaskyriazi2023realfusion} introduces a novel approach to single-image 3D reconstruction by leveraging pre-trained 2D diffusion generators for multi-view synthesis. The system employs specialized prompting to guide the diffusion model in generating plausible additional views, which are then integrated with neural radiance fields for complete object reconstruction.

\section{Task Setup and Data}

\subsection{Task Definition} 
Our work focuses on Novel View Synthesis within the context of image-to-3D generation. The core objective is to generate target view images given one or more conditioning images and their corresponding camera poses, defined by relative rotation and translation parameters. This task serves as a crucial component in the broader challenge of 3D content generation.

\subsection{Objective}
We aim to enhance Zero-1-to-3~\cite{liu2023zero1to3}, a conditional latent diffusion model that leverages image embeddings and camera pose information for novel view generation. Our analysis of the diffusion UNet's cross-attention module reveals that the original implementation's cross-attention effectively reduces to a linear transformation, limiting its capacity to process conditional information. Based on this finding, we propose architectural improvements to better utilize the cross-attention mechanism and enhance novel view synthesis quality.

\subsection{Dataset Configuration}
We utilize the Objaverse dataset~\cite{deitke2022objaverse}, following the preprocessing pipeline established in Zero-1-to-3. The dataset comprises over 800,000 3D models. 

\subsubsection{Processing Pipeline}
\begin{enumerate}
    \item \textbf{Camera Sampling}: For each object, 12 camera extrinsic matrices are randomly sampled, positioned to capture the object from varied viewpoints.
    
    \item \textbf{View Generation}: Each camera position generates a rendered view using ray-tracing, creating a comprehensive multi-view dataset per object.
    
    \item \textbf{Training Configuration}: Our models support two primary configurations:
    \begin{itemize}
        \item Base configuration: Two views per object (one target, one condition)
        \item Extended configuration: Variable number of conditioning views based on model architecture
    \end{itemize}
\end{enumerate}

\subsubsection{Dataset Scope}
Due to computational constraints, our experiments utilize a 25\% subset of the preprocessed Objaverse dataset. While this represents a reduced scale compared to the original Zero-1-to-3 training regime, it provides sufficient data to validate our architectural improvements and theoretical analysis.

\section{Technical Analysis of Cross-Attention in Zero-1-to-3}

\subsection{Initial Investigation of Conditional Latent Diffusion}
Cross-attention within the UNet architecture is fundamental to conditional latent diffusion models~\cite{Rombach_2022_CVPR}. In our initial analysis of Zero-1-to-3's architecture, we encountered an apparent paradox in the cross-attention mechanism. Specifically, when examining the tensor dimensions and disregarding batch size, we observed that the UNet's hidden state (a 2D matrix) cross-attends with a combined image and camera context embedding (a 1×768 1D vector) to produce the next hidden state.

For clarity of analysis, we focus on single-head attention. The query matrix Q is derived from the hidden state, while the key matrix K and value matrix V are derived from the context embedding. This configuration appeared problematic, as it suggested that attention weights would collapse to a 1D vector with uniform values—a significant limitation that motivated deeper investigation into the implementation.

\begin{figure}
    \centering
    \includegraphics[width=\linewidth]{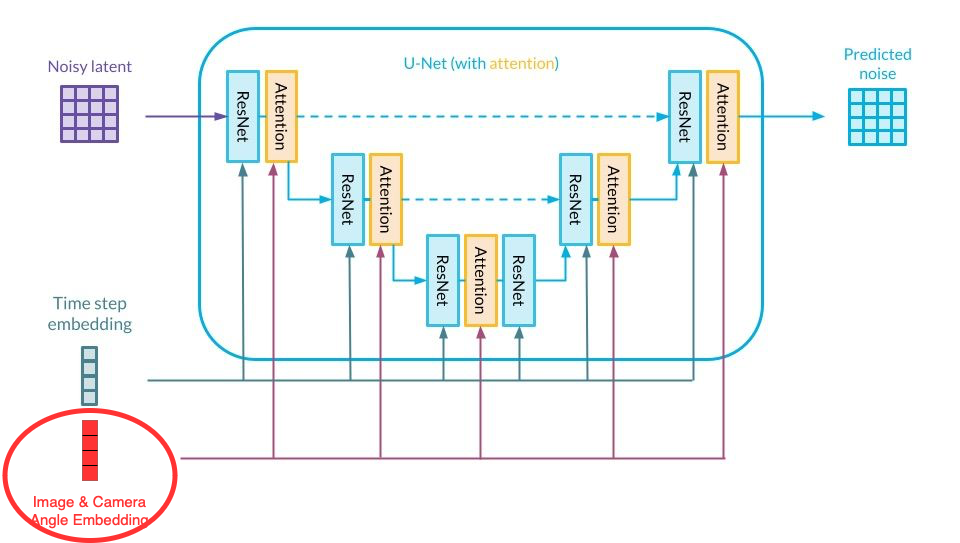}
    \caption{Cross-attention mechanism within a single U-Net layer during the diffusion process of Zero-1-to-3. The attention weights exhibit unexpected behavior due to architectural constraints.}
    \label{fig:zero123-unet}
\end{figure}

\subsection{Mathematical Analysis}
The context embedding is computed as:
\[ c := f(\text{CLIP}(I), P) \]
where \( I \) represents the input image and \( P = (r, \phi, \theta) \) denotes the camera position. The function \(f\) is a learnable linear transformation that maps the concatenated CLIP embedding and camera parameters to the context space.

The standard attention mechanism is defined as:
\[ \text{Attention}(Q, K, V) = \text{softmax}\left(\frac{QK^T}{\sqrt{d_k}}\right)V \]

With the softmax function:
\[ \sigma(\mathbf{z})_i = \frac{e^{z_i}}{\sum_{j=1}^{K} e^{z_j}} \]

Our analysis reveals that with only a single context vector available for attention, all attention weights default to one, i.e., $\text{Attention} = 1_{d\_\text{context}}$. This effectively reduces the cross-attention layer to duplicating the V vector multiple times, making the Q and K projections computationally redundant.

\subsection{Implementation Evidence}
We confirmed this limitation through code analysis, specifically identifying the issue in the Zero-1-to-3 codebase at \href{https://github.com/cvlab-columbia/zero123/blob/f426883b1a7353d91ddc34a551dd91b6223e4ce8/zero123/ldm/modules/attention.py#L189C9-L189C13}{line 189} where the attention tensor defaults to uniform values when context embedding is provided.

\subsection{Performance Analysis}
Despite this architectural limitation, Zero-1-to-3 achieves reasonable performance. This can be attributed to the residual connection in the Transformer module that combines the UNet hidden state with the processed context embedding \href{https://github.com/cvlab-columbia/zero123/blob/f426883b1a7353d91ddc34a551dd91b6223e4ce8/zero123/ldm/modules/attention.py#L215C9-L216C59}{as shown here}. However, this additive residual connection offers significantly less expressive power compared to proper cross-attention, suggesting substantial room for improvement in the model's architecture.

\section{Proposed Model One: Multi-view Zero-1-to-3}

\subsection{Motivation}
\begin{figure}
    \centering
    \includegraphics[width=1\linewidth]{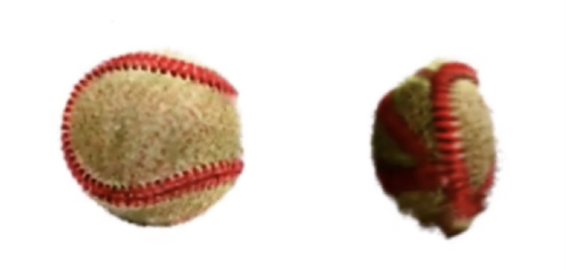}
    \caption{Example of back-view generation artifacts from RealFusion~\cite{melaskyriazi2023realfusion}, demonstrating the limitations of single-view conditioning.}
    \label{fig:back-view image}
\end{figure}

Single-view conditioning often results in degraded performance when generating back views of objects~\cite{qian2023magic123}, as illustrated in Figure \ref{fig:back-view image}. This limitation motivated our investigation into leveraging multiple input views with corresponding camera angles to provide richer spatial information and ensure view consistency.

\subsection{Architectural Design}
\begin{figure}
    \centering
    \includegraphics[width=1\linewidth]{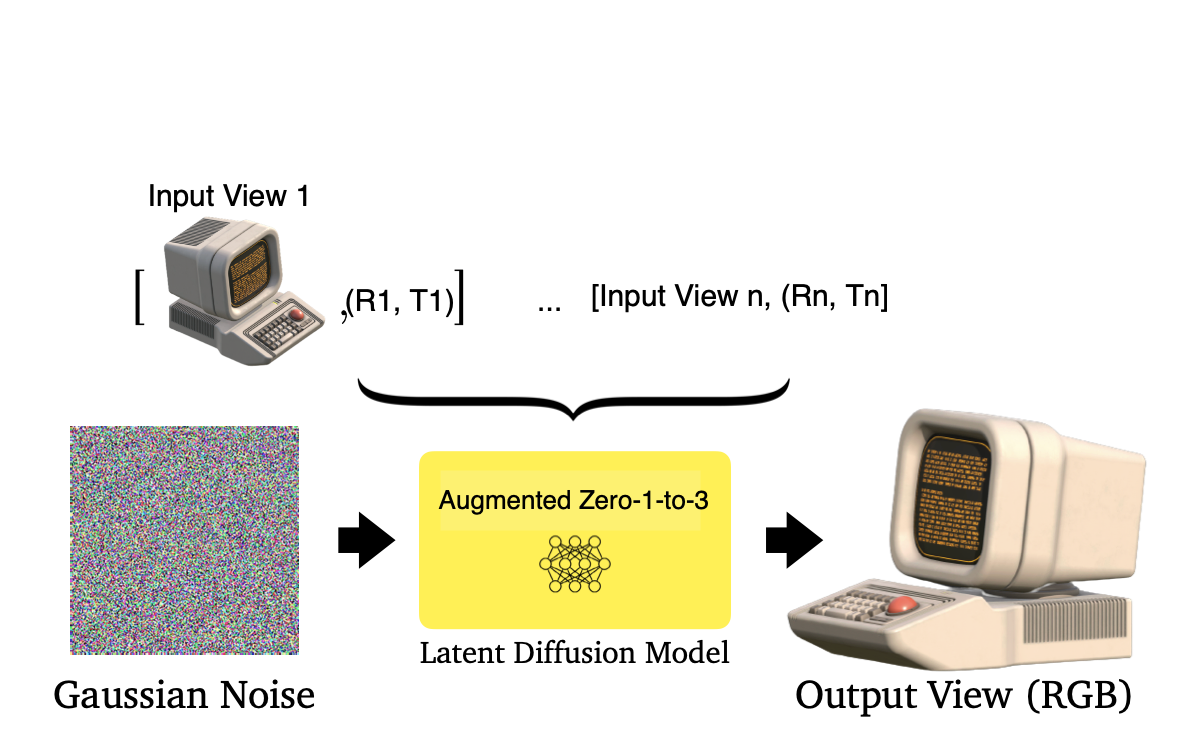}
    \caption{Multi-view architecture overview. Multiple input views and their camera poses are processed through parallel encoding paths before being combined for novel view generation.}
    \label{fig:multi-view-architecture}
\end{figure}

The architecture processes multiple input views through the following components:

\begin{enumerate}
    \item \textbf{View Encoding}: Each input view and its camera angle are independently encoded
    \item \textbf{Cross-Attention Integration}: Information from multiple views is combined through cross-attention mechanisms
    \item \textbf{Context Fusion}: View-specific features are aggregated into a unified representation
\end{enumerate}

\subsection{Training Methodology}

\subsubsection{Loss Function} \label{subsec:loss_function}
Given a dataset of paired images and camera views $\{x, x_{(R, T)}, R, T\}$, we extend the original Zero-1-to-3 denoising objective to incorporate multiple views:

\begin{align*}
    &\min_{\theta} \mathbf{E}_{t , \epsilon \sim \mathbf{N}(0,1)} 
    \bigg\| \epsilon - \epsilon_\theta\bigg(z_t, t, \\
    &\quad \text{Concat}[\{c(x_i, R_i, T_i)_i^n \}]\bigg)\bigg\|_2^2
\end{align*}

where the concatenated embedding $Concat[\{c(x_i, R_i, T_i)_i^n \}]$ combines information from n different views.

\subsubsection{Training Data Configuration}
From the Objaverse dataset, we sample N views per object from the available 12 rendered views:
\begin{enumerate}
    \item N-1 views serve as conditional inputs
    \item One view is designated as the target for training
    \item Views are randomly selected to ensure robust learning across different viewpoints
\end{enumerate}

\subsection{Hyperparameter Configuration}
The model's performance is controlled by three key hyperparameters:
\begin{enumerate}
    \item \textbf{View Count}: Number of conditional images and corresponding camera views (N)
    \item \textbf{Guidance Scale}: Controls the influence of classifier-free guidance
    \item \textbf{Diffusion Steps}: Number of denoising steps during inference
\end{enumerate}

\subsection{Implementation Status}
Due to computational constraints and the identified limitations in the original cross-attention mechanism, we have postponed full implementation of this architecture. Our analysis suggests that addressing the fundamental cross-attention issues should take precedence over scaling to multiple views. This led to our development of the second proposed model, which focuses on correcting the attention mechanism's limitations.

\subsection{Expected Outcomes}
While not fully implemented, theoretical analysis suggests this architecture would:
\begin{enumerate}
    \item Improve geometric consistency across generated views
    \item Reduce artifacts in occluded regions
    \item Scale performance with the number of conditional views
    \item Better handle complex object geometries
\end{enumerate}

\section{Proposed Model Two: Revamped embedding}
\subsection{Inspirations: Problem with cross-attention in Zero-1-to-3}
As mentioned before, we want to make sure the cross-attention could really work for diffusion process, so we redesign the embedding of image and angle condition by concatenating a image encoded vector and a camera angle vector. 

\begin{figure}
    \centering
    \includegraphics[width=1\linewidth]{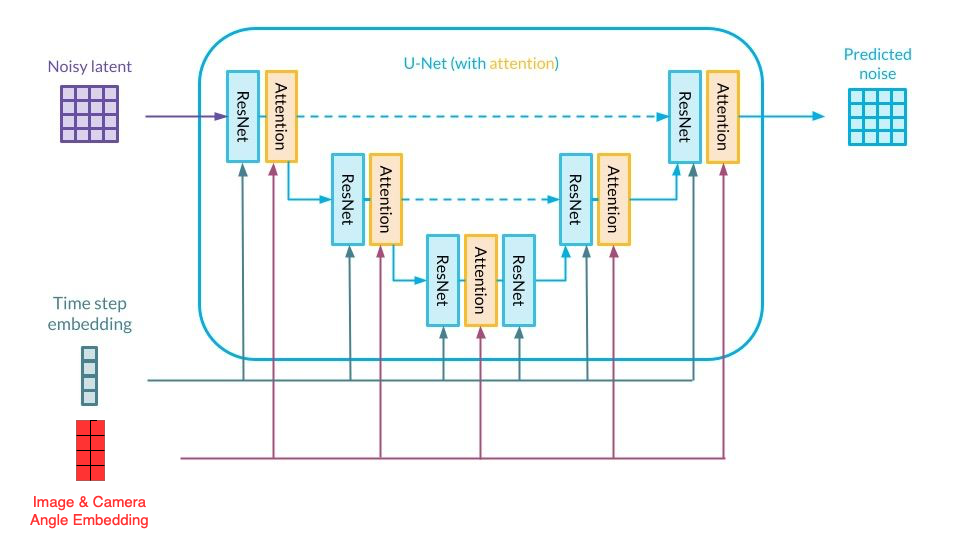}
    \caption{Architecture for our proposed fix to Zero-1-to-3 with separate positional information embedding. Image courtesy of Deepsense.ai}
    \label{fig:revamped-position-embedding}
\end{figure}

Zero-1-to-3 initially encodes the image condition and the angle condition separately, then concatenates them horizontally, and projects the combined condition into a lower dimension. In contrast, our methods directly encodes the image condition and the angle condition into the same dimension and concatenate them vertically, as illustrated in Figure \ref{fig:revamped-position-embedding}. Then the combined condition could be utilized by cross-attention mechanism to fully utilize the context provided by image and angle.

\begin{figure}
    \centering
    \includegraphics[width=1\linewidth]{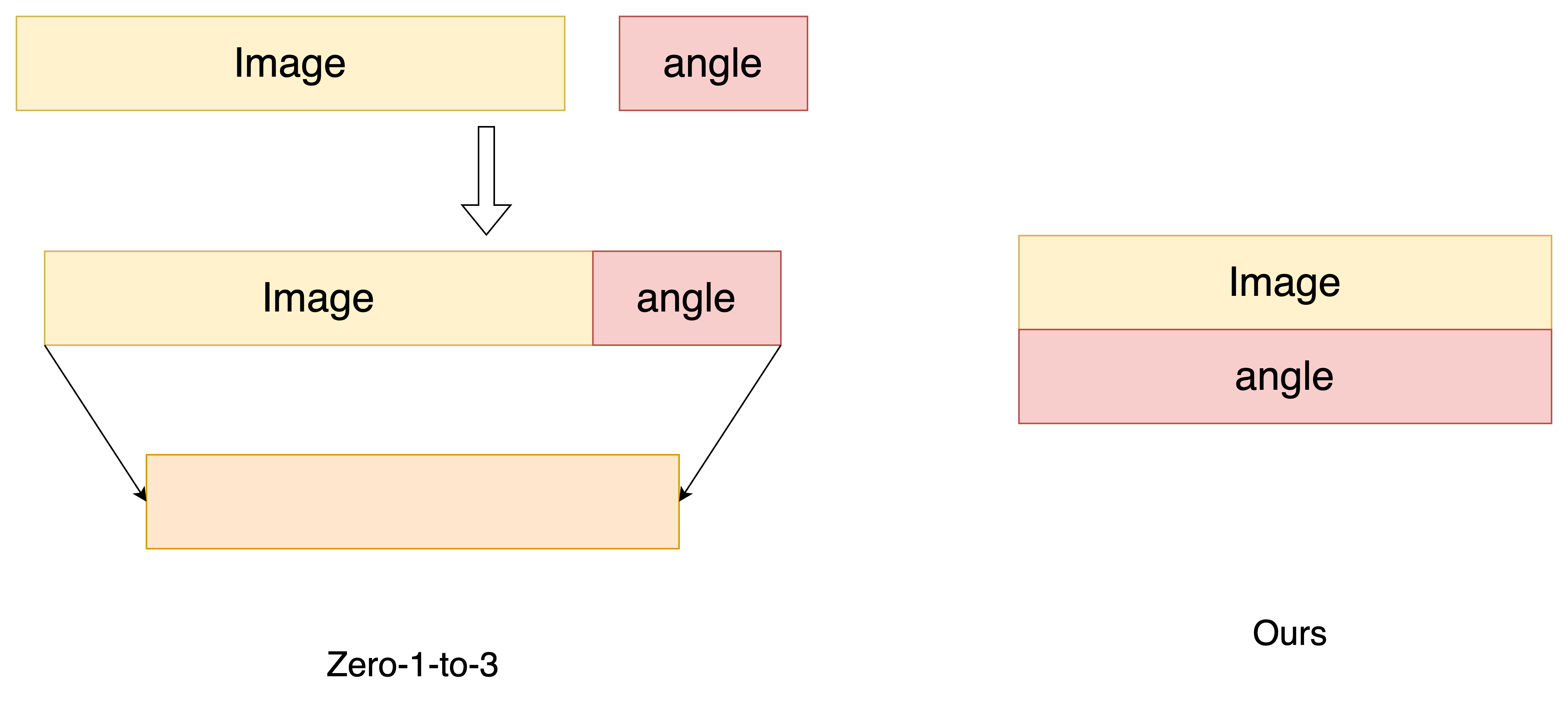}
    \caption{The embedding difference between Zero-1-to-3 and our revamped embedding.}
    \label{fig:embedding-dif}
\end{figure}



\subsubsection{Hyperparameters}
The hyperparameters could be projected dimension of combined image and angle information, diffusion guidance scale, and number of diffusion inference steps. The results have been listed below.

\subsubsection{Loss Function}
The loss function is the same as the loss function in \ref{subsec:loss_function}.

\subsection{Training Data}
As previously mentioned, we use the Objaverse dataset. Our methods use the same training data as Zero-1-to-3, we sample 2 views from 12 views for each object, then take one view as conditional view while designating another one view as the output target for training.

\subsection{Results and Analysis}

\begin{table*}[ht]
\centering
\label{tab:view-synthesis-comparison}
\begin{tabular}{lccc}
\toprule
Method & CLIP-Similarity $\uparrow$ & PSNR $\uparrow$ & LPIPS $\downarrow$ \\
\midrule
RealFusion \cite{melaskyriazi2023realfusion} & 0.735 & 20.216 & 0.197 \\
Magic123 \cite{qian2023magic123} & 0.747 & 25.637 & 0.062 \\
Zero-1-to-3 \cite{liu2023zero1to3} & 0.759 & 25.386 & 0.068 \\
\bottomrule
\end{tabular}
\caption{Quantitative comparison of baseline image-to-3D methods on novel view synthesis.}
\label{tab:effects-models}
\end{table*}

We conducted experiments with two sets of parameters: diffusion guidance scale (referred to as ``scale'' in the table) and the number of diffusion inference steps (referred to as ``steps'' in the table). The scale represents the magnitude of the diffusion condition's impact on the generation process, while the steps represent the number of iterations of the generation. The results for the input image of the Minions are summarized in Table \ref{tab:minions}.

However, due to limited time and GPU resource, we are unable to execute the same training steps as proposed by \cite{liu2023zero1to3}. Therefore, our proposed models have underwhelming generation quality, as displayed in the tavle. So, we will skip the metrics evaluation, which we leave for future work along with full fine-tuning of our proposed models.


\section{Intrinsic Metrics}
\begin{table*}[ht!]
\centering

\includegraphics[width=0.4\textwidth]{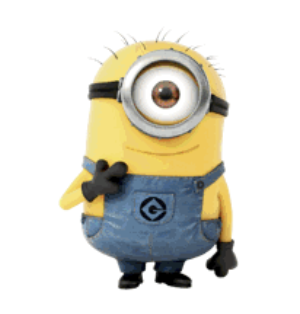} 
\vspace{5pt} 

\begin{tabular}{|p{2cm}|p{3cm}|p{3cm}|p{3cm}|p{3cm}|}
\hline
\diagbox{Step}{Scale} & \textbf{1} & \textbf{4} & \textbf{16} & \textbf{30} \\ \hline
20 &
\includegraphics[width=0.18\textwidth]{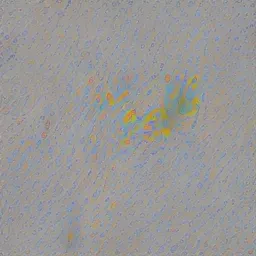} & 
\includegraphics[width=0.18\textwidth]{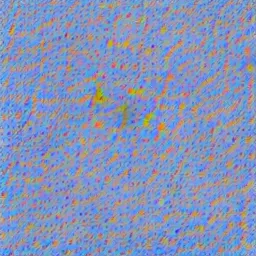} & 
\includegraphics[width=0.18\textwidth]{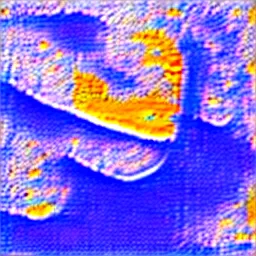} &
\includegraphics[width=0.18\textwidth]{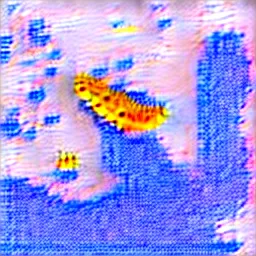} \\ \hline
50 & 
\includegraphics[width=0.18\textwidth]{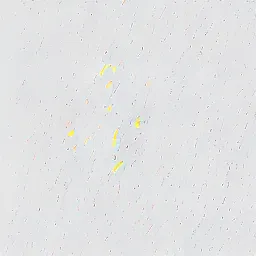} & 
\includegraphics[width=0.18\textwidth]{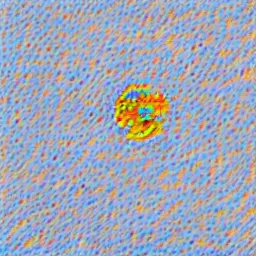} & 
\includegraphics[width=0.18\textwidth]{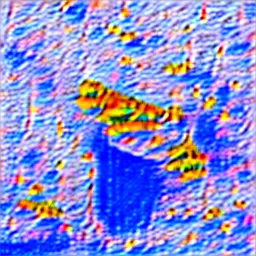} &
\includegraphics[width=0.18\textwidth]{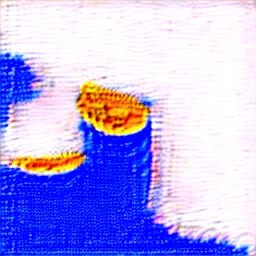} \\ \hline
100 & 
\includegraphics[width=0.18\textwidth]{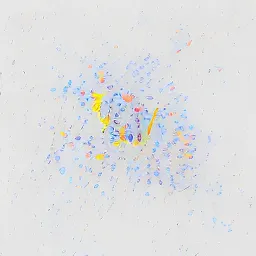}& 
\includegraphics[width=0.18\textwidth]{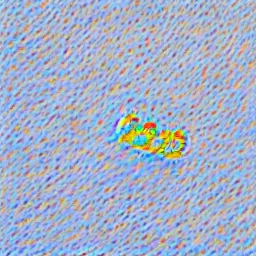}& 
\includegraphics[width=0.18\textwidth]{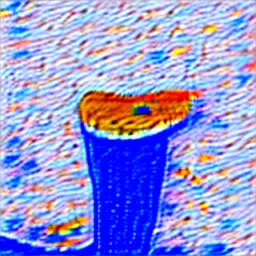}&
\includegraphics[width=0.18\textwidth]{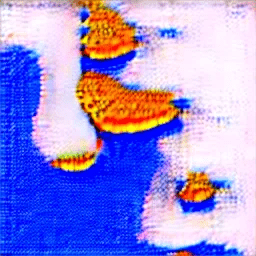} \\ \hline
150 & 
\includegraphics[width=0.18\textwidth]{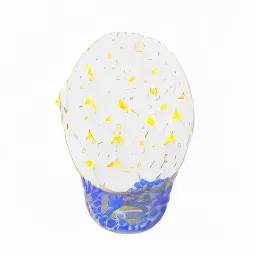}& 
\includegraphics[width=0.18\textwidth]{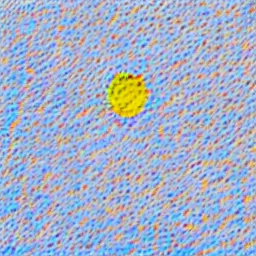}& 
\includegraphics[width=0.18\textwidth]{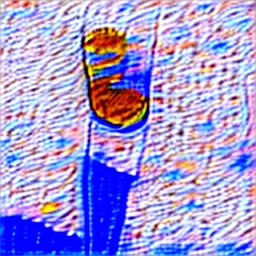}&
\includegraphics[width=0.18\textwidth]{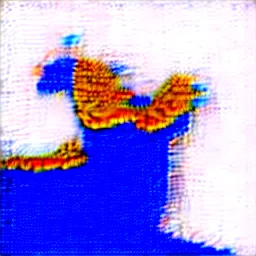} \\ \hline
200 & 
\includegraphics[width=0.18\textwidth]{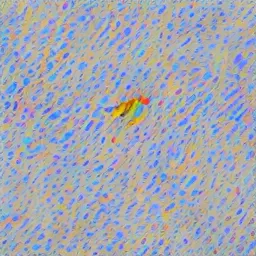}& 
\includegraphics[width=0.18\textwidth]{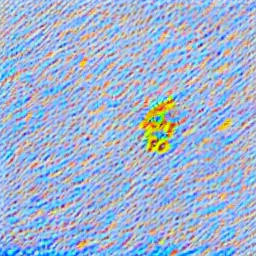}& 
\includegraphics[width=0.18\textwidth]{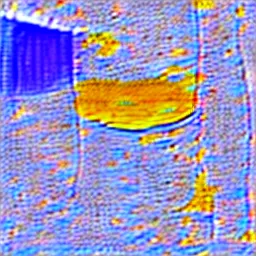}&
\includegraphics[width=0.18\textwidth]{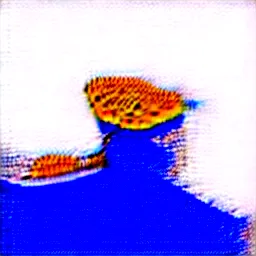} \\ \hline
\end{tabular}

\caption{Generative views for Minions based on our current model w.r.t diffusion steps and scale of the guidance constraints. Larger scale means larger influence from the classifier-free guidance.}
\label{tab:minions}
\end{table*}

In tasks involving novel-view synthesis, we want the novel view to be grounded on the conditional views and to be clear and sharp. The following intrinsic metrics captures these properties.

\subsection{Intrinsic Metric 1: FID}
FID (Frechet Inception Distance) \cite{heusel2018gans} computes the distance between the distribution of real and generated images. Given that our chosen model primarily generates novel viewpoint images before proceeding with 3D reconstruction, it is logical to assess the FID of the generated viewpoints relative to the original, as an indirect measure of the 3D object's quality and fidelity. 

\subsection{Intrinsic Metric 2: PSNR}
PSNR stands for Peak Signal-to-Noise Ratio. It is a metric used to measure the quality of a reconstructed image compared to its original version. Specifically, PSNR is a ratio between the maximum possible power of a signal (in this case, the original image) and the power of corrupting noise (the errors) that affects the fidelity of its representation (the reconstructed image). Similar to FID, PSNR is valuable in the 3D generation task as it offers a clear metric to gauge the fidelity of the newly generated views compared to the original model, highlighting the accuracy in detail and texture preservation.

\subsection{Intrinsic Metric 3: LPIPS}
LPIPS (Learned Perceptual Image Patch Similarity) \cite{zhang2018unreasonable} is a metric that leverages deep learning to assess the perceptual similarity between two images. In contrast to FID, which evaluates the distribution of features in generated images against real ones, LPIPS focuses more on how similar two specific images appear to human observers, simulated by neural networks. This perceptual approach makes LPIPS particularly useful for assessing 3D generation, where the key concern is not just the statistical accuracy of the generated images but also how realistically they replicate the original 3D model from various viewpoints.

\subsection{Intrinsic Metric 4: CLIP-similarity}
The novel-view rendering of the 3D object should ideally be very close to the real views despite angle differences. CLIP image embedding \cite{radford2021learning} allows us to compute the similarity of real views with generated views. Higher similarity indicates better model performance at creating the 3D object.

\subsection{Intrinsic Metric 5: CLIP-score}
Similar to the previous metric, we can also compare the CLIP similarity~\cite{hessel2021clipscore} between the text and image embeddings to capture the text-image alignment. 

\section{Future work and Limitations}
While our work addresses important limitations of Zero-1-to-3 and proposes effective improvements, there are still several potential directions for future research and some limitations to consider. One important direction for future work is to investigate the scalability of our approach to larger and more diverse datasets. While we have demonstrated the effectiveness of our method on a specific dataset, it would be valuable to assess its performance on a wider range of object categories, scene types, and image resolutions.

Another limitation of our model is the potential sensitivity of our method to the quality and consistency of the conditioning images. While our approach is designed to handle multiple views, it may still struggle if the input views are highly inconsistent or contain significant noise or artifacts.

Lastly, while our method can generate novel views, it may not always produce accurate results that respect real-world geometry, especially for objects with complex or ambiguous 3D structures. Incorporating additional geometric constraints and other complementary information, such as multi-view stereo or depth estimation, could help mitigate this issue. The issue may also come from the fact that our current model has not finished training or that the hyperparameters need to be further fine-tuned. Further experiments are needed to examine the capability of our model.

\section{Ethical Concerns and Considerations}
One potential concern of these models is the generation of misleading or deceptive content. As novel view synthesis methods become more capable of producing highly realistic images, there is a risk that they could be used to create fake or manipulated images. This could have serious consequences to society, where the spread of misinformation and misinformation is already a significant problem. In the future, more research are needed to align the outputs of 3D generative models with respect to human expectations and safety concerns.

\clearpage

\bibliographystyle{acl_natbib}
\bibliography{references}


\end{document}